\documentclass{article} 
\usepackage{template, times}

\usepackage{amsmath,amsfonts,bm}

\def\eqref#1{equation~\ref{#1}}

\def\1{\bm{1}}

\DeclareMathAlphabet{\mathsfit}{\encodingdefault}{\sfdefault}{m}{sl}
\SetMathAlphabet{\mathsfit}{bold}{\encodingdefault}{\sfdefault}{bx}{n}

\usepackage{url}
\usepackage{epstopdf}
\usepackage{hyperref}
\usepackage{adjustbox} 
\usepackage{mathtools, cuted}
\usepackage{natbib}
\usepackage{enumerate}
\usepackage{enumitem}
\usepackage{multirow}
\usepackage{caption}
\usepackage{subcaption}
\usepackage{enumitem}
\usepackage{tabularx}  
\usepackage{booktabs}  
\usepackage{bbding}
\usepackage{wasysym}
\usepackage{amssymb}
\usepackage{colortbl}
\usepackage{tcolorbox}
\definecolor{msblue}{RGB}{0,102,204}
\usepackage{tabularx}
\usepackage{wrapfig}
\newcommand{\ours}{$\text{TRAIT}$}

\renewcommand{\headrulewidth}{0pt}

\title{Task Oriented In-Domain Data Augmentation}

\author{
Xiao Liang$^{1,3}$\thanks{\noindent Equal contribution. Work done during Xiao Liang's internship at Microsoft Research Asia.}~,
Xinyu Hu$^{2*}$,
Simiao Zuo$^{2}$,
Yeyun Gong$^{3}$\thanks{\noindent Corresponding authors.}~,
Qiang Lou$^{2}$, 
Yi Liu$^{2}$,\\
\textbf{
Shao-Lun Huang$^{1}$,
Jian Jiao$^{2\dagger}$,
}
\\
~$^1$Tsinghua University
~$^2$Microsoft AI
~$^3$Microsoft Research
}

\iclrfinalcopy

\begin{document}
\maketitle

\begin{abstract}
\label{sec:abstract}
\noindent Large Language Models (LLMs) have shown superior performance in various applications and fields. 
To achieve better performance on specialized domains such as law and advertisement, LLMs are often continue pre-trained on in-domain data.
However, existing approaches suffer from two major issues. First, in-domain data are scarce compared to general domain-agnostic data. Second, data used for continual pre-training are not task-aware, such that they may not be helpful to downstream applications.
We propose TRAIT, a task-oriented in-domain data augmentation framework. Our framework is divided into two parts: in-domain data selection and task-oriented synthetic passage generation. The data selection strategy identifies and selects a large amount of in-domain data from general corpora, and thus significantly enriches domain knowledge in the continual pre-training data. The synthetic passages contain guidance on how to use domain knowledge to answer questions about downstream tasks. By training on such passages, the model aligns with the need of downstream applications.
We adapt LLMs to two domains: advertisement and math. On average, TRAIT improves LLM performance by 8\% in the advertisement domain and 7.5\% in the math domain.
\end{abstract}
\section{Introduction}
\label{sec:intro}

Large language models (LLMs) have achieved significant performance improvements in various applications such as language modeling \citep{brown2020language, touvron2023llama, chowdhery2023palm} and visual understanding \citep{radford2021learning}. They have also shown superior performance in fields such as finance \citep{xie2023efficient}, e-commerce \citep{ma2023ecomgpt} and healthcare \citep{bakhshandeh2023benchmarking}. However, the models are usually trained on a large amount of general domain-agnostic data, such as web corpora. Because of the lack of domain-specific training, LLMs suffer from subpar performance when directly applied to certain domains such as advertisement.

To adapt LLMs to a specific domain, continual pre-training methods \citep{gururangan2020don} are commonly applied. In particular, the LLM is continual pre-trained on in-domain corpora, such that it can acquire domain knowledge and better adapt to downstream tasks. Existing works \citep{cheng2023adapting} have shown that such a technique drastically improves performance of LLMs on domains such as law and bio-medicine.

There are two major issues when continual pre-training LLMs. First, in-domain data are scarce. LLMs are pre-trained on large domain-agnostic corpora. For example, the web corpus used for pre-training contains more than ten trillion tokens. However, domain-specific data are magnitudes smaller, i.e., the ads in-domain corpus in our experiments contains only several billion tokens. Such a data scarcity issue significantly hinders model performance after continual pre-training.

Second, in-domain data used for continual pre-training are not task-oriented. Many existing works \citep{achiam2023gpt, li2023textbooks, liu2024best, shao2024deepseekmath} focus on generating or selecting in-domain data without considering the downstream tasks. That is, the continual pre-training data are often passages that describe keywords/concepts of the target domain, which are generated or selected without considering whether the passages benefit downstream tasks.

We propose {\ours} (\underline{\textbf{T}}ask o\underline{\textbf{R}}iented in-dom\underline{\textbf{AI}}n data augmen\underline{\textbf{T}}ation), a data augmentation framework driven by downstream tasks of the domain. The framework is divided into two parts.
First, to address the data scarcity issue of in-domain corpora, we propose a data selection strategy. The proposed algorithm identifies in-domain data from general corpora, and also applies a quality filter to ensure that the selected data have high educational value \citep{gunasekar2023textbooks}. In practice, the amount of selected data is magnitudes larger than the in-domain dataset. For example, for the advertisement applications, the in-domain dataset contains about 1B tokens, and {\ours} selects an additional 15B tokens from web corpora.

Second, we propose a task-oriented synthetic passage generation guideline. Specifically, each generated passage contains multiple problems, where each problem comes from a different downstream task from the domain. Then, for each problem, {\ours} generates a problem-specific paragraph that suggests possible answers to the problem. Additionally, the synthetic passage also contains an \emph{enlightenment paragraph}. This paragraph focuses on relationships among problems in the passage, including common and individual characteristics that are used to generate answers. Intuitively, the problem-specific paragraphs teach the model how to use techniques to solve a particular problem. And the enlightenment paragraph teaches the model common and unique aspects of all problems in the domain.

To fully exploit the power of {\ours}, we employ a two-stage training strategy for continual pre-training.
In the first stage, the model is trained with in-domain data, including both the original in-domain corpora and the selected data. In this stage, the model adapts to the domain by learning domain knowledge.
Then, in the second stage, the model is trained with the task-oriented passages. During this stage, the model learns how to use domain knowledge to solve problems, such that it better aligns with the need of downstream tasks.

We conduct extensive experiments by adapting LLMs to two domains: advertisement and math. For the advertisement domain we consider 7 downstream tasks, and {\ours} improves existing continual pre-training methods by 6.5\% average score, while improving the base LLM (without continual pre-training) by 8\%.
For the math domain we consider 9 downstream tasks, where {\ours} outperforms the baseline by 5.5\% average score and outperforms the base LLM by 7.5\%. For the challenging MATH task in math domain, {\ours} outperforms the base LLM by over 15\%.

\section{Method}
\label{sec:method}

\subsection{Overview}

We propose {\ours}, a data augmentation framework with two components.
First, we propose a data selection strategy to select in-domain data from general corpus. In this way, we can significantly enlarge the domain-specific training data, such that the data contain more domain knowledge compared with the original small in-domain corpus.
Second, we propose a guideline to generate task-oriented passages from in-domain data. The synthetic passages focus on using domain knowledge to solve the given tasks.

We use both the in-domain data and the synthetic passages to continual pre-train LLMs. Specifically, we first train the model on in-domain data, such that the model can learn more domain knowledge. Then, we train the model on the task-oriented synthetic passages. During this stage, the model learns to solve downstream tasks using the acquired domain knowledge. The proposed data augmentation framework and training strategy can drastically improve model performance by adapting LLMs to specific domains.

\subsection{In-Domain Data Selection}
\label{sec:augmentation}
In practice, the size of general corpus is orders of magnitude larger than domain-specific corpus. For example, the ads domain corpus contains about 1B tokens in our experiments, while the general web corpus contains trillions of tokens. To alleviate such a data scarcity issue, we propose to select in-domain data from general corpus.

We train a FastText \citep{joulin2017bag} classifier to identify in-domain data from large amount of domain agnostic data. 
Specifically, to train the FastText classifier, we select a certain number of in-domain data as positive samples and the same amount of out-of-domain data as negative samples. The trained binary classifier is then used to select in-domain data from the general corpus (e.g., the web corpus).

We apply a filter to ensure that the in-domain data (both the original in-domain corpus and the selected data) have high educational value \citep{gunasekar2023textbooks}. In this way, we can boost the quality of the filtered in-domain data, which in turn improves performance of the models.

The proposed data selection strategy has two benefits. First, it can significantly enrich in-domain data.
In practice, the amount of selected data is magnitudes larger than the in-domain dataset. For example, the original ads domain corpus contains about 1B tokens in our experiments, and we select an additional 15B tokens from the web corpus (after selection and filtering). 
Second, the data selection strategy enables \textit{replay} \citep{ibrahim2024simple}, such that generality of LLMs is largely kept after continual pre-training (see Table~\ref{tab:general_results} for experiments). In more details, for a specific LLM, replay happens when the continual pre-training data contain a certain amount of pre-training data (e.g., the web corpus). It has been observed that replay is crucial to keep LLM's generality (e.g., instruction following) after training.

\begin{figure*}[t]
\centering
\includegraphics[width=1\linewidth]{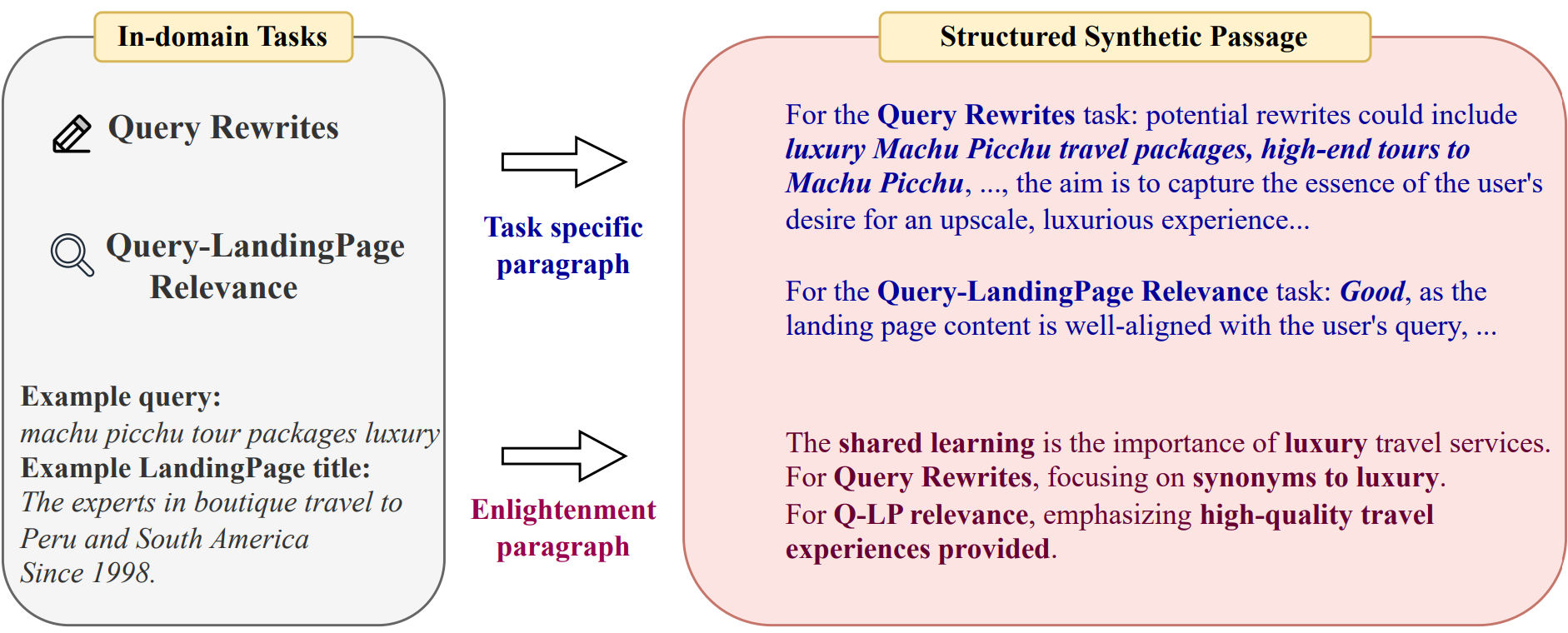}
\caption{An example of a task-oriented synthetic passage on the ads domain. Left: two downstream tasks (Query Rewriting and Query-LandingPage Relevance) and inputs. Right: the structure of the generated passage, including two problem-specific paragraphs and an \emph{enlightenment} paragraph.} 
\label{fig:ad_passage}
\end{figure*}

\subsection{Synthetic Data Generation}

To adapt LLMs to a specific domain, we first train the model on in-domain data, such that the model can acquire domain knowledge. Another key aspect crucial to model performance is the model's ability to use such knowledge to solve downstream tasks. To address this issue, we propose a guideline to generate task-oriented synthetic passages. In this way, the model is aware of downstream tasks during continual training, and thus model performance can be significantly improved.
In the next section, we describe how to generate the task-oriented passages in detail.

\section{Task-Oriented Passage Generation}
\label{sec:synthetic}

\subsection{Guideline}

The goal of the synthetic passages is to describe how to solve downstream tasks using domain knowledge. 
In {\ours}, each synthetic passage describes the common and individual characteristics of all downstream tasks in a domain. This resembles human learning: we learn how to solve individual problems, as well as common knowledge that can be applied to all problems.

We propose a guideline to generate task-oriented synthetic passages:
\begin{itemize}[leftmargin=10pt]
    \item[$\diamond$] We build each passage using several problems, where each problem comes from a different downstream task.
    \item[$\diamond$] Within a passage, for each problem we generate a problem-specific paragraph that suggests possible answers to the problem. Different prompts are used to generate paragraphs for different tasks, while the same prompt is used for problems from the same task.
    \item[$\diamond$] For each passage, we generate an \emph{enlightenment paragraph}. This paragraph emphasizes relationships among problems, including shared and individual characteristics that are used to generate answers to the problems.
\end{itemize}

The enlightenment paragraph requires summarizing common and individual aspects of different problems from different downstream tasks. This is natural for certain domains. For example, in the ads domain in Figure~\ref{fig:ad_passage}, different questions in the passage ask about different aspects of the \emph{same query}. As another example, in the finance domain, different features of the \emph{same company} may be useful for different tasks. We call these domains \emph{entity-centered}. These domains focus on \textbf{understanding of entities} from various aspects.

On the other hand, in domains such as math and physics, the common aspects of different problems are not entities, but knowledge or techniques that can be applied to solve the problems.
For example, in the math domain, each passage may contain multiple questions that require different techniques to solve, e.g., the GSM8k benchmark focuses on simple arithmetic, while the MATH benchmark focuses on logical reasoning.
We call these domains \emph{knowledge-centered}. These domains focus on using \textbf{universal knowledge} to solve problems.

\begin{figure*}[t]
\centering
\includegraphics[width=1\linewidth]{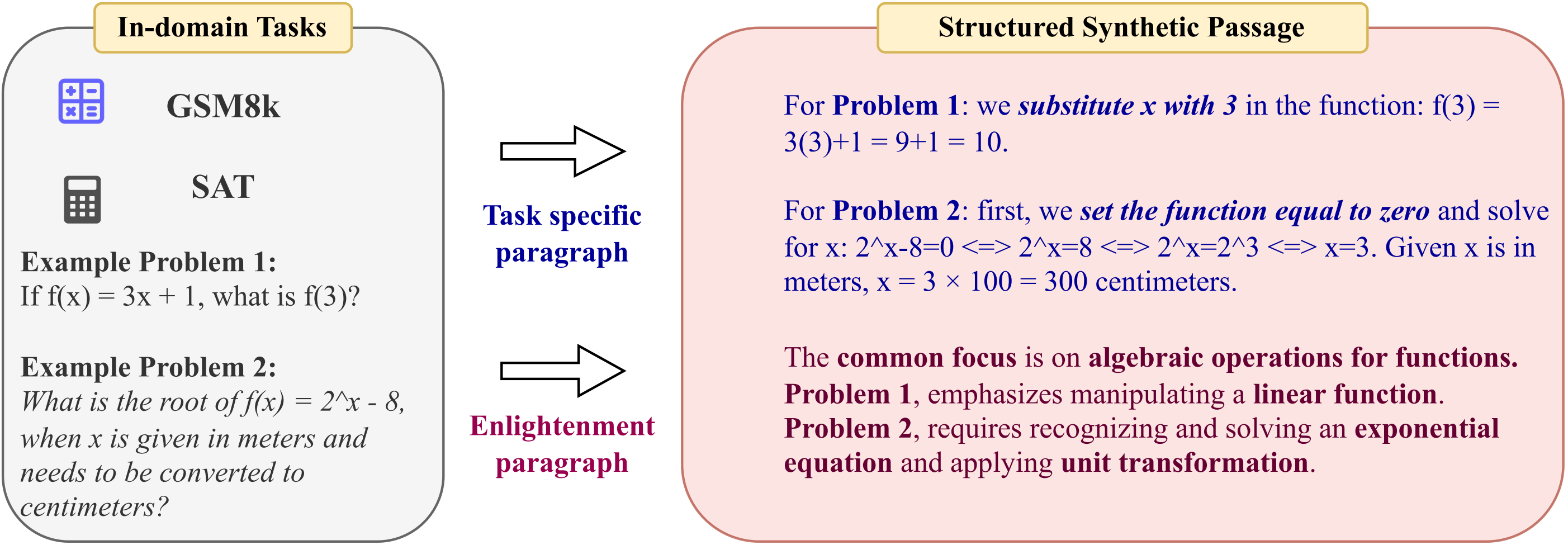}
\caption{An example of a task-oriented synthetic passage on the math domain. Left: the selected two tasks (GSM8k and SAT) with an example problem from each task. Right: the structure of the generated passage, including two problem-specific paragraphs and an \emph{enlightenment}
paragraph.}
\label{fig:math_passage}
\end{figure*}

\subsection{Example: Task-Oriented Passage for Entity-Centered Domains}

Recall that each synthetic passage is divided into two parts: problem-specific paragraphs and an enlightenment paragraph. We use ads domain as an example to illustrate how the two components are generated.

We select problems from different downstream tasks in the ads domain. In Figure~\ref{fig:ad_passage}, the passage contains two tasks (or two problems): Query Rewriting and Query-LandingPage Relevance. 

For Query Rewriting, the task is to generate variations that maintain the search intent but diversify the expression. For the query "\textit{Machu Picchu tour packages luxury}", the generated problem-specific paragraph looks like: \textit{Potential rewrites for the query could include ``luxury Machu Picchu travel package'', ``high-end tours to Machu Picch''.}

For Query-LandingPage Relevance, the task is to decide whether the content of the landing page (the webpage to which the user is directed after clicking on an ad) addresses the intent of the search query. For the query "\textit{Machu Picchu tour packages luxury}" and the landing page "\textit{The experts in boutique travel ...}", the generated problem-specific paragraph looks like: \textit{The landing page details, such as the expertise of the travel specialists ... directly correspond to the user’s search for a luxurious Machu Picchu tour, demonstrating a strong relevance.}

We also generate an enlightenment paragraph that focuses on relationships among the downstream tasks. For example, in Figure~\ref{fig:ad_passage}, the enlightenment paragraph is: \textit{The shared learning is the importance of the {luxury and personalized aspects of the travel service}. For Query Rewrites, focusing on {synonyms} related to luxury and high-end services. Ensuring high relevance in Query-Landing Page Relevance emphasizes {the high-quality travel experiences to Machu Picchu}.}

The enlightenment paragraph severs as a central tenant of intelligence. It demonstrates which aspects of the entity are needed in all downstream tasks, and which aspects are tailored for a specific task. Such an explicit signal significantly boosts model performance. For example, for some downstream tasks in the ads domain, adding the enlightenment paragraph brings a 3\% performance gain (see Table~\ref{tab:ablation_on_2stage_enlightment} for details).

\subsection{Example: Task-Oriented Passage for Knowledge-Centered Domains}

Different from entity-centered domains such as ads and finance, in knowledge-centered domains such as math and physics, the focus is on applying universal knowledge to solve problems. We use math domain as an example to demonstrate how we generate the task-oriented passages.

In Figure~\ref{fig:math_passage}, we select two problems from the GSM8k and the SAT benchmarks. Then, we generate problem-specific paragraphs to solve the problems. Similar to the ads domain, in the enlightenment paragraph we summarize the common and individual techniques that are used to solve the problems.
In more details, the enlightenment paragraph states that the common knowledge used to solve the two problems is "\textit{algebraic operation for functions}". Each task requires additional techniques: 
the GSM8k problem requires "\textit{manipulating a linear function}, while the SAT problem requires "\textit{solving exponential equation}".

We remark that in knowledge-centered domains, the model learns universal knowledge that can be applied to all problems, i.e., the techniques learned from one problem are applicable to other problems.
Therefore, the purpose of the enlightenment paragraph is to communicate about the techniques required to solve problems. The paragraph explicitly points out the common techniques needed by all downstream tasks, such that the model gains better awareness of the importance of such techniques. The model also learns task-specific techniques as pointed out by the enlightenment paragraph.

\section{Data Preparation}
\label{sec:data}

We apply our data augmentation method, \ours, to two domains: advertisement (ads) and math. In this section, we detail the process for in-domain data selection and synthetic passage generation. Refer to Appendix.~\ref{sec:examples} for examples and more details.

\subsection{Task-Oriented Passage Generation}
Following the guidelines for task-oriented passage generation, we select problems from each downstream task and use GPT-4 to generate the full passage (prompt details can be found in Appendix.~\ref{sec:examples}).
This approach is adaptable to any new target domain, leveraging GPT-4's understanding of various domains and its ability to handle diverse tasks within them. By selecting relevant problems and utilizing our generation prompt, our method ensures effective application across multiple domains.

\subsection{In-Domain Data Selection}
\label{sec:ads_data}
\textbf{Ads domain.}
We train a domain-specific FastText classifier for in-domain data selection, as detailed in Section~\ref{sec:augmentation}. First, we randomly select 500k positive samples from the ads in-domain corpus.
We also select 500k negative samples from Slimpajama \citep{cerebras2023slimpajama}, alpaca \citep{alpaca}, OpenHermes-2.5 \citep{OpenHermes-2.5} and Tulu-v2 \citep{ivison2023camels}. 
Then, for model training, we set the model dimension to 256, learning rate to 0.1, the maximum word n-gram length to 3, the minimum word occurrence to 3, and the epoch to 3.
Next, we apply the trained classifier to select samples with the highest scores from fineweb \citep{penedo2023refinedweb}.
Finally, we apply a quality filter to the data to ensure that each sample has an educational value over 1.5 \citep{gunasekar2023textbooks}, where in total we select 15B filtered tokens.

\noindent \textbf{Math domain.}
The domain-specific FastText classifier is trained similar to that in the ads domain. For the positive samples, we sample 200k examples from open-source benchmarks (such as GSM8k and SAT) and 200k samples from OpenWebMath \citep{paster2023openwebmath}. The negative samples are constructed similar with the ads domain.
Due to the scarcity of math-related content in the general corpus, we retrieved the math data from a combination of MathPile \citep{wang2023mathpile} and Proof-Pile-2 \citep{azerbayev2023llemma}, resulting in a collection of around 5.5 billion tokens.

\section{Experiments}
\label{sec:setup}

\begin{table*}[t]
\centering
\small
\begin{tabular}
{l|cc|cc|cccc|c}
\toprule

\multirow{2}{*}{Method} & \textbf{QAC} & \textbf{QLP} & \multicolumn{2}{c|}{\textbf{QR}} & \textbf{AG} & \textbf{DG} & \textbf{TG} & \textbf{TR} & \multirow{2}{*}{\textbf{Avg. $\triangle$}} \\

& \multicolumn{2}{c|}{Auc} & Den. & \multicolumn{1}{c|}{Div.} & \multicolumn{4}{c|}{Win Rate (\%)} & \\

\midrule
\multicolumn{10}{c}{\texttt{Few-shot Results}} \\
\midrule
Mistral-7B &  \textbf{69.48} & 59.54 & -- & -- & -- & -- & -- & -- & --  \\

Random & 63.94 & 60.29 & -- & -- & 42.20 & 55.15 & 53.62 & 47.75 & -1.54\% \\

DSIR   & 60.03 & 60.27 & -- & -- & \textbf{57.43} & 50.18 & 50.88 & 51.95 & +1.41\% \\

\rowcolor{gray!20}
TRAIT & 65.18 & \textbf{65.91} & -- & -- & 51.55 & \textbf{60.42} & \textbf{54.60} & \textbf{55.10} & \textbf{+7.97\%} \\

\midrule
\multicolumn{10}{c}{\texttt{Fine-tuned Zero-shot Results}} \\
\midrule

Mistral-7B &  82.93 & 78.81 & 5.29 & 3.06 & -- & -- & -- & -- & -- \\

Random    & 83.44 & 78.83 & 5.45 & 3.29 & 46.02 & \textbf{52.35} & 50.22 & 48.55 & +0.68\%  \\

DSIR    & 84.10 & 79.96 & 5.48 & 3.36 & 50.58 & 49.38 & 51.98 & 50.32 & +2.60\%  \\

\rowcolor{gray!20}
TRAIT &  \textbf{84.40} & \textbf{80.71} & \textbf{5.57} & \textbf{3.36} & \textbf{50.15} & 51.95 & \textbf{52.98} & \textbf{54.43} & \textbf{+4.79\%} \\

\bottomrule
\end{tabular}
\caption{Evaluation results of downstream tasks in the ads domain. Here, \textit{Avg $\triangle$} is the average relative improvement over all evaluation metrics for all tasks. Best results highlighted in \textbf{bold}.}
\label{tab:main_results}
\end{table*}

\begin{table*}[t]
\centering
\small
\adjustbox{max width=0.98\linewidth}{
\begin{tabular}{l|ccccccccc|c}
\toprule
Method  & 
\textbf{GSM8K}  & 
\textbf{MATH$^\dagger$} & 
\textbf{SVAMP}  & 
\textbf{ASDiv}  & 
\textbf{MAWPS}  & 
\textbf{TAB}    & 
\textbf{MQA}    & 
\multicolumn{1}{c}{\textbf{\begin{tabular}[c]{@{}c@{}}MMLU\\ STEM\end{tabular}}} & \textbf{SAT} & 
\textbf{Avg.} \\
\midrule

Base   & 40.9 & 12.4 & 65.4 & 68.5 & 87.4 & 52.7 & 34.6 & 49.3 & 65.6 & 53.0 \\
Random & 34.8 & 14.0 & 60.4 & 65.2 & 82.4 & 39.7 & 34.9 & 46.4 & 56.2 & 48.2 \\
DSIR   & 46.4 & 22.4 & 64.5 & 72.7 & 88.0 & 47.1 & 38.6 & 43.2 & 71.9 & 55.0 \\
\rowcolor{gray!20}
TRAIT      & \textbf{56.4} & \textbf{28.0} & \textbf{71.8} & \textbf{76.0} & \textbf{89.5} & \textbf{53.1} & \textbf{46.1} & \textbf{49.5} & \textbf{75.0} & \textbf{60.5}\\

\bottomrule
\end{tabular}
}
\caption{Few-shot CoT reasoning results of downstream tasks in the math domain. For MATH$^\dagger$, evaluation is performed on OpenAI’s MATH subset~\citep{lightman2023let}, as the original test samples may be included in public training sets. Best results highlighted in \textbf{bold}.}
\label{tab:math-results}
\end{table*}

We evaluate {\ours} by adapting LLMs to the ads and math domains via continual pre-training. 
In all the experiments, we use \href{https://huggingface.co/mistralai/Mistral-7B-v0.1}{Mistral-7B} \citep{jiang2023mistral} as the base model.
We compare {\ours} with two data selection baselines: (1) \textit{Random sampling}, which randomly selects samples from open-source general corpora; and (2) \textit{DSIR} \citep{xie2023data}, an importance sampling strategy for selecting in-domain data from general corpora, such that the selected data distibutionally similar with the in-domain data.
To promote fair comparisons, all models (including the base Mistral-7B model, baseline models, and {\ours}) are trained on the same amount of data with the same computational budget.
Details about the training process can be found in Appendix \ref{sec:training_details}.

\subsection{Ads Domain}
\label{sec:ads_exp}

\textbf{Downstream tasks}.
We consider seven tasks within the ads domain. There are two classification tasks: Query-AdCopy Relevance (\textbf{QAC}) examines the relevance of a user's query to the ad copy, while Query-LandingPage Relevance (\textbf{QLP}) assesses relevance between a user's query and the advertisement's landing page content. 
For generation tasks, we focus on generating dynamic content: Query Rewriting (\textbf{QR}) generates rewrites of user queries, Ad Copy Generation (\textbf{AG}) creates complete ad copy directly, and both Description Generation (\textbf{DG}) and Title Generation (\textbf{TG}) develop concise descriptions and titles from the ad's landing page information. Additionally, Title Rewriting (\textbf{TR}) enhances user engagement by refining the ad title to better suit the user's query and the original title context.

\noindent \textbf{Evaluation settings.}
We evaluate {\ours} under both few-shot and fine-tuning settings. Each task contains 5k test samples. For the fine-tuning setting, each task contains 30k training samples.

$\diamond$ For the two natural language understanding tasks (QAC and QLP), we adopt Area under curve (Auc) as the evaluation metric. 

$\diamond$
For AG, DG, TG and TR, we use ChatGPT \citep{OpenAIchatgpt2022} to calculate the winning rate of {\ours} compared with the Mistral-7B model. Specifically, we prompt ChatGPT to choose the better answer from responses generated by our model and Mistral-7B. In order to mitigate ChatGPT's positional bias for evaluation \citep{chen2024humans}, we swap the positions of the two responses and prompt ChatGPT again to choose the better answer. We average the outcomes from the two rounds as the final winning rate.

$\diamond$
The evaluation metrics for QR consist of diversity (Div.) and density (Den.), with details provided in Appendix~\ref{sec:eval_queryrewrites}.

\noindent \textbf{Results.}
Experimental results are summarized in Table~\ref{tab:main_results}. 
From the results, we see that {\ours} significantly outperforms both the Mistral-7B model and the baselines. 
Specifically, {\ours} achieves average increases of 8.0\% and 4.8\% across all downstream tasks compared with Mistral-7B in the few-shot and fine-tuning settings, respectively.
And the proposed framework outperforms the best performing baseline by 6.5\% and and 2.2\% in the few-shot and fine-tuning settings, respectively.
In the few-shot setting, {\ours} outperforms all the baselines in 4/6 tasks; while in the fine-tuning setting, the proposed framework performs the best in 6/7 tasks.

\begin{table*}[t]
\centering
\small
\begin{tabular}
{l|ccccc|cccc}
\toprule
& \multicolumn{5}{c|}{\texttt{Ads Domain}}
& \multicolumn{4}{c}{\texttt{Math Domain}}
\\
& \textbf{QAC} 
& \textbf{QLP}
& \textbf{DG} 
& \textbf{TG} 
& \textbf{TR} 
& \textbf{GSM8k}  
& \textbf{SAT} 
& \textbf{MAWPS}     
& \textbf{MATH$^\dagger$}   \\

\midrule

{\ours} &  
84.40 &  80.71 &  51.95 &  52.98 &  54.43
&  56.4 &  75.0 &  89.5  &  28.0 \\

w/o E.P.
&  83.77  &  79.80 &  50.95 &  51.92  &  51.58 
&  54.6 &  68.8 &  89.7 &  27.2  \\

w/o two-stage
&  83.23  &  79.84 &  50.22 &  51.50 &  51.32 
&  53.2 &  59.4 &  89.3  &  25.2 \\


\bottomrule
\end{tabular}
\caption{
Effectiveness of the enlightenment paragraph and the two-stage training approach. We adopt the fine-tuning settings for the ads domain and the few-shot settings for the math domain. Here \textit{w/o E.P.} means the model is trained without the enlightenment paragraphs.}
\label{tab:ablation_on_2stage_enlightment}
\vskip -0.1in
\end{table*}

\subsection{Math Domain}

\textbf{Downstream tasks.}
We evaluate the models across nine mainstream benchmarks: GSM8k \citep{cobbe2021training}, MATH \citep{hendrycks2021measuring}, SVAMP \citep{patel2021nlp}, ASDIV \citep{miao2021diverse},  MAWPS \citep{koncel2016mawps}, TabMWP (TAB) \citep{lu2022dynamic}, MathQA (MQA) \citep{amini2019mathqa}, MMLU-STEM \citep{hendrycks2020measuring}, and SAT \citep{azerbayev2023llemma}. For evaluation, we adopt the math evaluation suite\footnote{\href{https://github.com/ZubinGou/math-evaluation-harness}{math-evaluation-harness}}. 

\noindent \textbf{Evaluation settings.} For all tasks, we evaluate under the few-shot chain-of-thought (CoT) \citep{wei2022chain} setting. We use accuracy as the final evaluation metric.

\noindent \textbf{Results.}
As shown in Table~\ref{tab:math-results}, our continual pre-trained model achieves an absolute average accuracy improvement of 7.5\% across all benchmarks compared with Mistral-7B, with a significant gain of 15.6\% on the most challenging MATH benchmark. We remark that {\ours} outperforms all the baselines in all the tasks.

\label{sec:math_exp}

\subsection{Analysis}
\label{sec:results}

\begin{figure*}[t]
    \centering
    \begin{subfigure}[b]{0.48\textwidth}
        \centering
        \includegraphics[width=\textwidth]{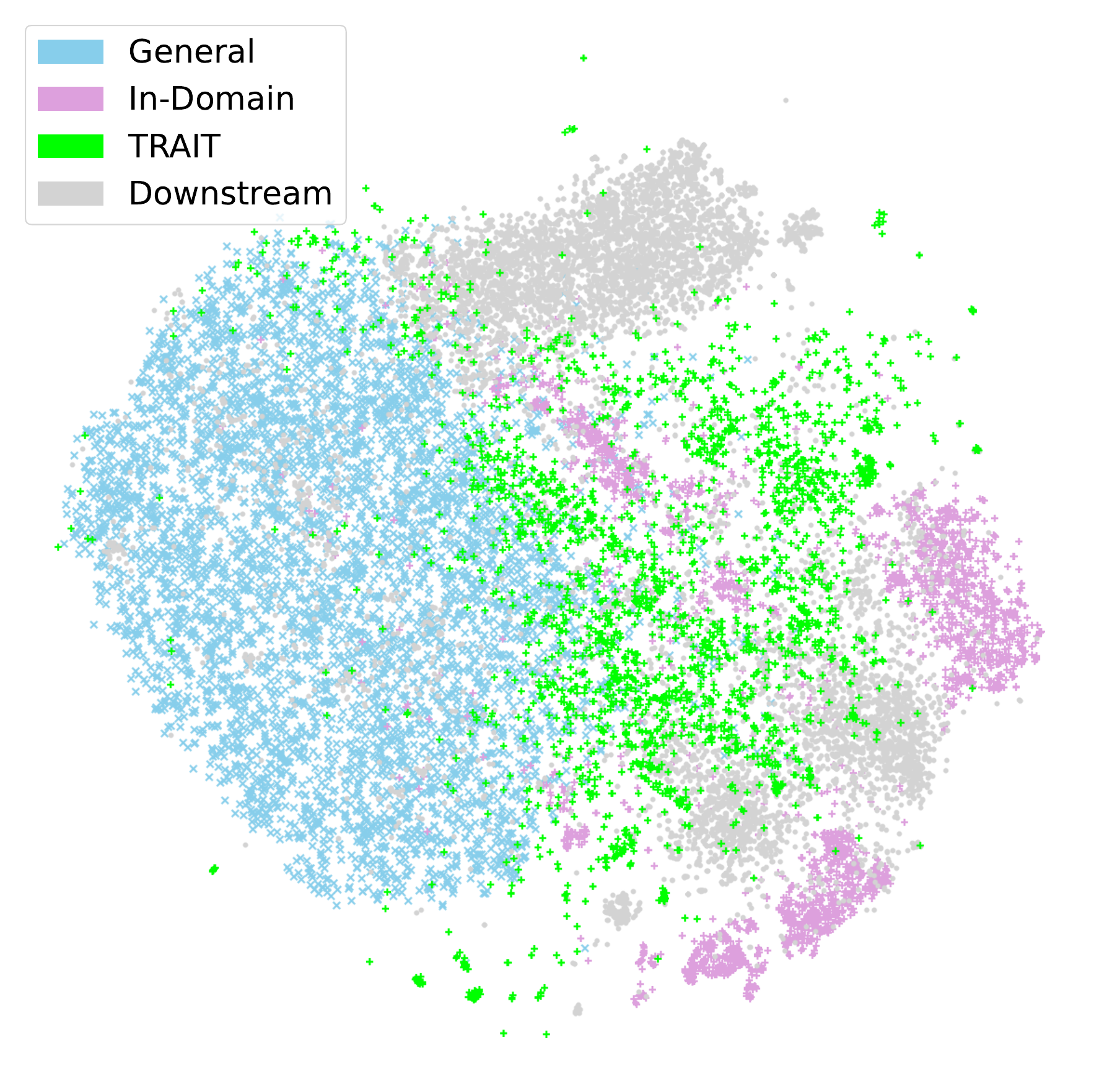}
    \end{subfigure}
    \hfill
    \begin{subfigure}[b]{0.48\textwidth}
        \centering
        \includegraphics[width=\textwidth]{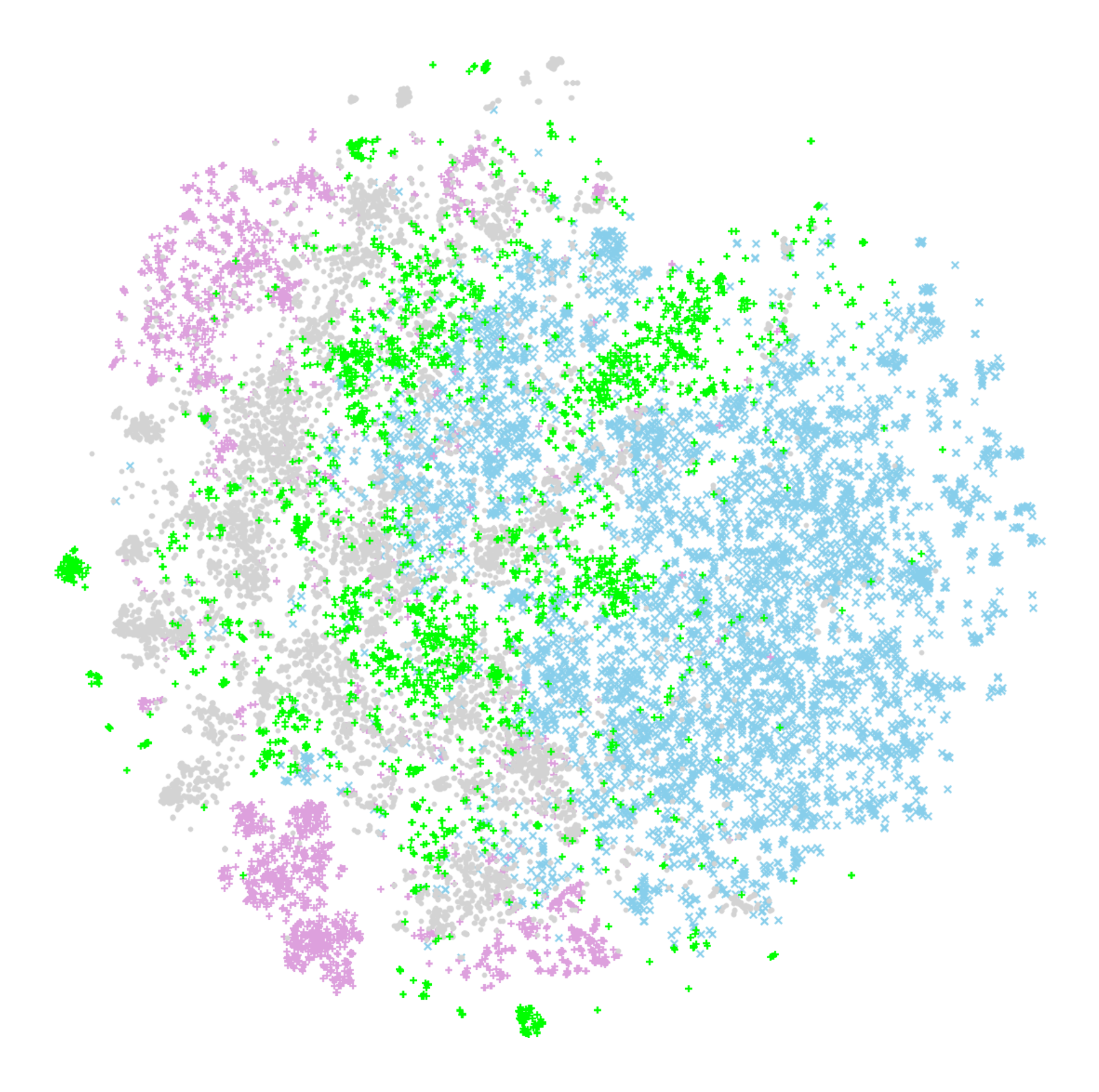}
    \end{subfigure}
    \vspace{-0.4cm}
    \caption{Visualization of samples from the general corpus, the original in-domain ads corpus, ads downstream tasks, and {\ours} (including both selected in-domain data and synthetic passages). We use Spacy~\citep{honnibal2017spacy} (left) and Mistral-7B~\citep{jiang2023mistral} (right) for embedding, while using t-SNE~\citep{van2008visualizing} for visualization.}
    \label{fig:embedding_tsne}
\end{figure*}

\textbf{Effectiveness of {\ours}.} In Figure~\ref{fig:embedding_tsne}, we see that the general data is far from the original in-domain data, indicating the necessity for domain-adaptive continual pretraining. The downstream tasks are distributed in various clusters, in proximity to the in-domain data, but not fully covered by it. For {\ours}, the mix of selected in-domain data and synthetic passages perfectly aligns with the downstream tasks, reflecting the task-awareness nature of our approach.

\begin{table}[t]
\centering
\small
\begin{tabular}{ccc|ccccc}
\toprule

\textbf{In-D.} & \textbf{Sel.} & \textbf{Syn.} & \textbf{QAC} & \textbf{QLP} & \textbf{TG} & \textbf{TR} \\

\midrule

\multicolumn{3}{c|}{Mistral-7B} & 82.93 & 78.81 & -- & -- \\ \midrule

\CheckmarkBold &  &  & 82.50 & 79.18 & 50.73 & 50.38 \\

& \CheckmarkBold & & 82.47 & 79.35 & 52.05 & 48.05 \\

\CheckmarkBold &  & \CheckmarkBold  & 82.64 & 79.17 & 52.88 & 48.20\\

& \CheckmarkBold & \CheckmarkBold  & 83.33 & 80.21 & 50.85 & 50.00 \\

\CheckmarkBold & \CheckmarkBold & \CheckmarkBold  & \textbf{84.40} & \textbf{80.71} & \textbf{52.98} & \textbf{54.43} \\

\bottomrule
\end{tabular}
\caption{Performance of models continual pre-trained on different data. Models are evaluated on the ads domain under the fine-tuning setting. Here, \textit{In-D.} means the original in-domain corpus, \textit{Sel.} means selected in-domain data, and \textit{Syn.} means synthetic passages.}
\label{tab:ablation_on_data_type}
\end{table}

In Table~\ref{tab:ablation_on_data_type}, we see the effect of each data component. The second row confirms the benefit of the original in-domain data, showing an average 1\% performance gain across all tasks. A more notable contribution comes from {\ours} augmented data, with a nearly 5\% gain observed, showing effectiveness of our data augmentation strategy.

Moreover, the benefit of the enlightenment paragraph is significant, as shown in Table~\ref{tab:ablation_on_2stage_enlightment}. It reinforces a deeper understanding of queries in the ads domain and focuses on shared problem-solving techniques in the math domain.

\begin{figure}[t]
    \centering
    \begin{subfigure}[b]{0.36\textwidth}
        \centering
        \includegraphics[width=\textwidth]{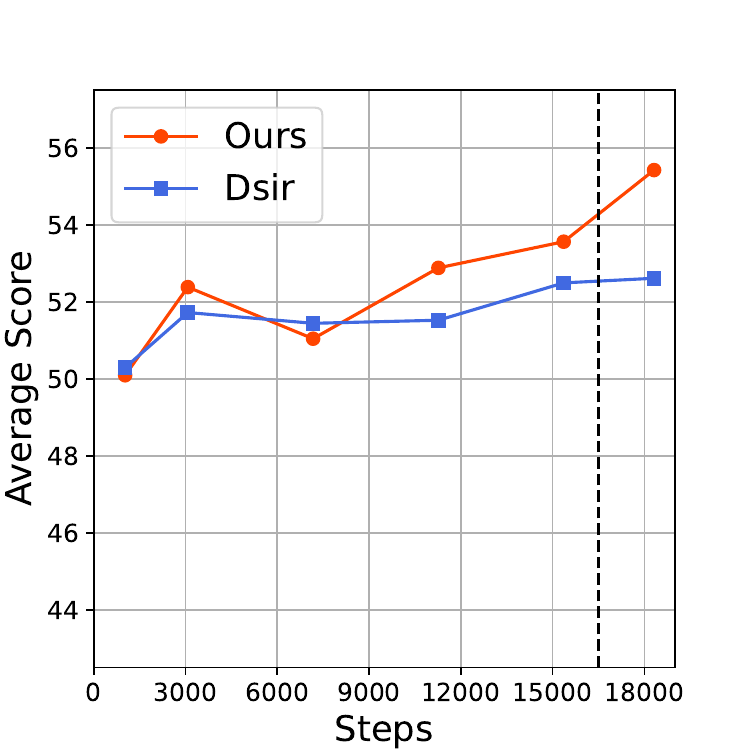}
    \end{subfigure}
    \hspace{0.15\textwidth}
    \begin{subfigure}[b]{0.36\textwidth}
        \centering
        \includegraphics[width=\textwidth]{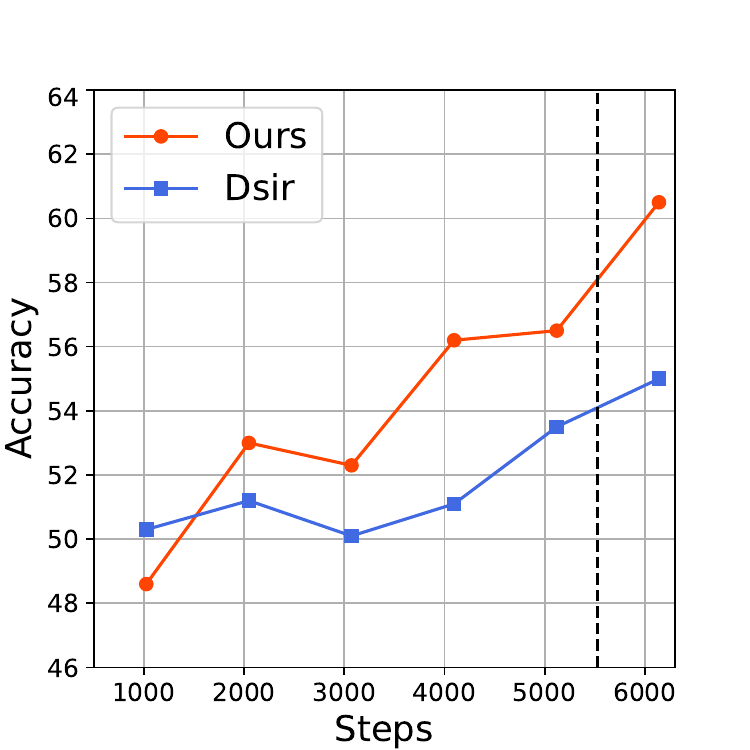}
    \end{subfigure}
    \caption{Left: The average winning rate of 4 ads generation tasks (AG, DG, TG and TR) during continual pre-training. Right: The average few-shot accuracy of all math tasks during continual pre-training.}
    \label{fig:performance_during_training}
\end{figure}

\noindent \textbf{Two-stage vs. Single-stage training.}
The performance of downstream tasks during continual pre-training is documented in Figure~\ref{fig:performance_during_training}. In the first stage, where the aim is to learn new in-domain knowledge, we observe fluctuations in downstream performance as new knowledge, which may not be directly relevant to the tasks, is acquired. In the second stage, the focus shifts to applying the learned knowledge to solve downstream tasks directly, resulting in a larger upward trend in task improvement. The overall benefit of adopting the two-stage training compared to a mixed single stage is significant, as shown in Table~\ref{tab:ablation_on_2stage_enlightment}.

\begin{table}[t]  
\centering
\small
\begin{tabular}{l|cccc}  
\toprule
  &   \textbf{BBH} &   \textbf{ARC}  &   \textbf{HellaSwag} &   \textbf{AgiEval} \\  

\midrule
Mistral-7B &   55.91 & 70.55 &    61.25 &   32.64 \\  \midrule

In-Domain  &   49.84 & 65.24  &   59.57 &   30.43 \\  

{\ours} & 53.06  & 67.59 &   60.34 &   32.26 \\

\bottomrule

\end{tabular}  
\caption{Few-shot evaluation of models trained on different data on general benchmarks. Here, \textit{In-Domain} means the model is continual pre-trained on ads in-domain corpus (without selected data).}
\label{tab:general_results}  
\end{table}  






\noindent \textbf{Generality after continual pre-training.}
Using the original in-domain data, the generality of LLMs deteriorates significantly, as shown in Table \ref{tab:general_results}, with the model's performance on BBH decreasing from 55.91 to 49.84. In contrast, the model using {\ours} retains much of the generality. This is because, during the first stage, we train on the selected in-domain data from the general corpus as a knowledge replay \citep{ibrahim2024simple} and refocus on the target domain. Additionally, high-quality synthetic data ensures the model is trained with extensive reasoning.

\section{Related Work}
\label{sec:related}
\subsection{Data Augmentation for Language Models}

Data selection is essential for the effective training of LLMs, as it significantly influences their performance. Common data selection methods include heuristic-based quality filters \citep{together2023redpajama, soldaini2024dolma}, lightweight classifiers \citep{joulin2017bag, brown2020language}, and perplexity (PPL)-based models \citep{heafield2011kenlm, wenzek2019ccnet}, which are often developed using curated sources such as Wikipedia. For domain-specific data, techniques usually involve extracting information from the open web using heuristics or applying specialized classifiers to ensure relevance \citep{ma2023ecomgpt, xie2023efficient}. 
Other approaches select data based on their added value compared to typical examples from the target domain, or employ n-gram hash models to identify samples related to that domain \citep{axelrod2017cynical, feng2022automatic, xie2023data}.

The use of synthetic data is becoming a key strategy for augmenting the training of LLMs, particularly useful in areas like mathematical reasoning \citep{gou2023tora,huang2024key,toshniwal2024openmath,li2024common} and general instruction following \citep{wang2022self,xu2023wizardlm,li2024synthetic}. The Phi series highlights the effectiveness of models trained solely on ``textbook quality'' synthetic data \citep{gunasekar2023textbooks, li2023textbooks, abdin2024phi}.

\subsection{Continual Pre-Training of LLMs}
Continual pre-training is increasingly recognized as an effective way to adapt large language models (LLMs) incrementally to new data or changes in domain focus without complete retraining. This method ensures the continuous integration of new knowledge, maintaining the model's relevance and effectiveness \citep{jin2021lifelong, loureiro2022timelms}.
\citet{gururangan2020don} has shown that continual pre-training can significantly improve model performance across various domains.
LLMs like EcomGPT and FinPythia demonstrate the application of continual pre-training in e-commerce and finance, using data from the open web and Common Crawl to stay functional and relevant \citep{ma2023ecomgpt, xie2023efficient}. 

\section{Conclusion}
\label{sec:conclusion}
This paper presents {\ours}, a task-oriented in-domain data augmentation framework for continual pre-training of large language models. 
The framework is divided into two parts. First, we select in-domain data from general domain-agnostic corpora to augment the training set. The augmented in-domain training corpus contain rich domain knowledge. 
Second, we generate task-oriented synthetic passages. These passages contain guidance on how to apply domain knowledge to answer questions about downstream tasks. 
We conduct extensive experiments by adapting LLMs to the advertisement and math domains. Experimental results validate the effectiveness of the proposed framework. Specifically, on average, {\ours} improves the base LLM (without continual pre-training) by over 5\% on both domains.

\bibliography{main}
\bibliographystyle{template}

\newpage
\appendix
\appendix

\section{Training Details}
\label{sec:training_details}
In this study, we select Mistral-7B~\citep{jiang2023mistral} as our base model for domain-adaptive continual pre-training. 
We perform continual pre-training using DeepSpeed~\citep{rasley2020deepspeed} with its ZeRO stage-1 optimizer, setting the batch size to 1M tokens and utilizing bf16 precision.
We adopt a Warmup-Stable-Decay (WSD) learning rate scheduler~\citep{hu2024minicpm} with a maximum learning rate of 2e-5, involving a 3\% warm-up phase and an exponential decay phase during the last 10\% of the training process.

During fine-tuning, we train all the models on downstream tasks for 5 epochs, using a cosine learning rate schedule with a maximum rate of 5e-6 and a 3\% warm-up phase. For both few-shot and fine-tuned settings, we use vLLM (0.4.2)~\citep{kwon2023efficient} to accelerate inference.

\section{Potential Risks of Training Solely on In-Domain Corpus}
Training a model solely on an in-domain corpus that lacks strict filtering for quality and diversity can lead to reduced effectiveness. While the original in-domain corpus helps to maintain relevance, its quality may not match that of carefully curated pre-training corpora used in leading open-source language models, such as Mistral-7B.~\citep{jiang2023mistral}. 


As demonstrated in Figure~\ref{fig:failure_case}, using only our specific in-domain corpus for training adversely affects the model’s ability to perform well in downstream generation tasks. Instead of providing accurate responses, it produces repetitive and meaningless text. In contrast, the base model, not subjected to this limited training, correctly generates appropriate responses.

However, by incorporating high-quality data from both retrieved and synthetic sources during training, we can preserve the model's original capabilities. This approach aligns with findings~\citep{ibrahim2024simple} that suggest replay techniques during continual training help maintain a model’s pre-trained skills.

\setlength{\columnsep}{0.2cm}
\begin{figure*}[ht]\footnotesize
\begin{tcolorbox}[colback=msblue!5!white,colframe=msblue!80!black]

\#\#\# \textbf{Prompt}: 

You are an expert in advertisement and your task is to generate a creative ad given its content.

Here is the content of the advertisement:

DocumentTitle: Flow Meters \& Controllers | Ferntree Gully, VIC | Anri §
Paragraph: Florite manufacture a range of high quality microcomputer-
based controllers suitable for any commercial, industrial or research
application such as: ; Flow Control – Power supply/readout/control units for
Mass Flow Meters (MFM’s)MFC’ programmable measurement units. Panel
mount or bench top. Also suitable for Pressure, temperature, level \& any
process control. call our team on 03 9752 4057. § OtherPrimaryContent: en/
mfd ;mcmflow. § VisualTitle: Reliable Flow/ Controllers for Any Industry §
MetaDescription: Controls supply flow meters \& controllers for businesses
Australia-wide. Call 03 9752 4057 or click here to learn more. §

The generated creative ad is:

\vspace{0.5\baselineskip}

\#\#\# \textbf{Base Model}: 

Flow Control: The Ultimate in Flow Metering. Experience the Elegance.

\vspace{0.5\baselineskip}

\#\#\# \textbf{Model Continual Trained on In-domain Unlabeled Data}: 

5' x 2' 3' x 8' 4' 6' x 10' x 12' 7' x 16' 9' x 13' x 14' x 15' x 18' x 20' x 23' x 24' x
27' x 29' x 30' x 40' x 50' 11' x 26' x 28' x 32' x 36' x 44' 17' x 60' 22' x 45' x 70'
x 75' x 80' x 90' x 105' x 110' x 120' x 130' x 100' x 140' x 160' x 180' x 200' x
220' x 240' x 270' x 300' x 190' x 250' x 280' x 260' x 310'

\vspace{0.5\baselineskip}

\#\#\# \textbf{Model Trained on Our Curated Corpus}: 

100\% Same-Day Delivery. Gophr: The Future of Last Mile Delivery.

\end{tcolorbox}
\caption{An example of model degradation: the model, when continually pretrained on the original in-domain corpus, exhibits repetitive and nonsensical text generation, unlike its performance with the base model and TRAIT corpus, where this issue is absent.}
\label{fig:failure_case}
\end{figure*}

\section{Evaluation Details of Query Rewriting}
\label{sec:eval_queryrewrites}
We generate ten query rewrites for each query and apply a quality filter to score the rewrites as either \textit{Good} or \textit{Bad}, based on their alignment with the original user query's intent. Density (Den) is defined as the total average number of good query rewrites per query. For diversity (Div), we use an off-the-shelf clustering algorithm to determine the average number of clusters per query based on all query rewrites. Both metrics range from 0 to 10.

\section{Examples of Prompts and Training Data}
\label{sec:examples}
In this section, we demonstrate examples of prompts we used along with all unlabeled and downstream fine-tuning data. 
Specifically, we present an example of the prompt for synthetic data in Figure~\ref{fig:synthetic_prompt}, 
an in-domain ad sample in Figure~\ref{fig:ad_example},
a synthetic ad sample in Figure~\ref{fig:example_of_synthetic_ads}, 
a synthetic math sample in Figure~\ref{fig:example_of_synthetic_math}, 
retrieved ad and math samples in Figure~\ref{fig:retrieved_ads} and~\ref{fig:retrieved_math} ,
and the downstream fine-tuning data for the ads domain in Figure~\ref{fig:prompt_qac}, \ref{fig:prompt_qlp}, \ref{fig:prompt_qr}, \ref{fig:prompt_ag}, \ref{fig:prompt_dg}, \ref{fig:prompt_tg} and \ref{fig:prompt_tr}.

\setlength{\columnsep}{0.2cm}


\begin{figure*}[ht]\footnotesize
\begin{tcolorbox}[colback=msblue!5!white,colframe=msblue!80!black]

\begin{center}
    \textbf{Prompt for Generating Task-Oriented Synthetic Data}
\end{center}

\#\#\#\# Structured Guideline for Passage Generation

\vspace{\baselineskip}

\#\#\#\# Inputs Required:

- **Questions**: The question for each task.

\vspace{\baselineskip}

\#\#\#\# Passage Generation Steps:

- Task specific: For each of the downstream tasks listed below, write one paragraph analyzing the potential answers and the reasoning process associated with each. Please list the answer explicitly.

- Enlightenment: After writing paragraphs for all tasks, highlighting shared learnings across all tasks and distinct problem solving tricks for each task, specifically the current problem.

\vspace{\baselineskip}
 
\#\#\#\# Quality Considerations:

- Ensure coherence and logical flow throughout the passage.

- Maintain a concise and clear writing style, avoiding redundancy and focusing on summarizing key points.

\vspace{\baselineskip}

\#\#\#\# Input:
Please return only the generated passage between tags <Passage></Passage> given below input.

- \{Task 1\}: \{Problem 1\}

- \{Task 2\}: \{Problem 2\}

- \{Task 3\}: \{Problem 3\}

- \{Task 4\}: \{Problem 4\}

\vspace{\baselineskip}

\end{tcolorbox}
\caption{The prompt utilized for generating the task-oriented synthetic data.}
\label{fig:synthetic_prompt}
\end{figure*}

\begin{figure*}[ht]\footnotesize
\begin{tcolorbox}[colback=msblue!5!white,colframe=msblue!80!black]

\begin{center}
    \textbf{Example for an unlabeled ad sample}
\end{center}

\textbf{Advertisement Title}: Vision Reading Glasse

\vspace{\baselineskip}

\textbf{Advertisement Description}: Save On Vision Reading Glasses. Everyday Low Prices!

\vspace{\baselineskip}

\textbf{Key words}: hand held lighted magnifying glass

\vspace{\baselineskip}

\textbf{Document Title}: Hand Held Lighted Magnifying Glass - Walmart.com

\vspace{\baselineskip}

\textbf{Heading}: Results for "Hand Held Lighted Magnifying Glass" (1000+) Options

\vspace{\baselineskip}

\textbf{Primary Content}: Departments Brand Speed Subscription Availability Special Offers Customer Rating Features Magnification Color Material
Manufacturer Part Number Count Retailer Gifting Price when purchased online Best seller Sponsored Magnifying Glass with 18 LED Light,
Meromore 30x Handheld Magnifier for Reading Save with Shipping, arrives in 2 days Meromore Magnifying Glass, Lighted Magnifying Glass with 3
LED, Handheld Magnifier with 3x 45x Magnification for Kids Reading Meromore 30x Magnifying Glass with 18 LED Lights, Black MagniPros 3 Ultra
Bright LED Lights 3X 4.5X 25X Power Handheld Reading Magnifying Glass with Light- Ideal for Reading Small Prints, Map, Coins, Inspection and
Jewelry Loupe Free shipping, arrives in 3+ days Magnifying Glass with Light, Lighted Magnifying Glass, 5X Handheld Pocket Magnifier Small
Illuminated Folding Hand Held Lighted Magnifier for Reading Coins Hobby Travel - 45 Mm Diameter 2 options KEKOY Handheld Magnifying Glass
with Light and Double Lens, Ultra Clear USB Charging Magnifying Glass for Close Work with 6x 9x 15x Detachable Lens, Strong Maginifier for
Reading RockDaMic Professional Magnifying Glass with Light (3X / 45x) Large Lighted Handheld Glass Magnifier Lupa for Reading, Jewelry, Coins,
Stamps, Fine Print - Strongest Magnify for Kids \& Seniors 2 sizes Large 4.35in Magnifying Glass 10X 35X with Light and Stand, Handheld Standing
LED Illuminated Magnifier, Folding Reading Magnifying Glass with for Seniors Read, Cross Stitch, Map, Jewelry Meromore Magnifying Glass.

\end{tcolorbox}
\caption{An example of an unlabeled in-domain sample from the ads domain.}
\label{fig:ad_example}
\end{figure*}

\setlength{\columnsep}{0.2cm}
\begin{figure*}[t]\footnotesize
\begin{tcolorbox}[colback=msblue!5!white,colframe=msblue!80!black]

\begin{center}
\textbf{Example for Syhthetic Ads Data}
\end{center}
For the Query Rewrites (QR) task, the potential answers revolve around creating variations of the user query "machu picchu tour packages luxury." Possible rewrites could include "luxury Machu Picchu travel packages," "high-end Machu Picchu tours," and "premium tour packages to Machu Picchu." Each rewrite aims to capture the essence of the user's request while incorporating keywords that might improve the ad's visibility and relevance. The reasoning process involves ensuring that the rewrites maintain the focus on luxury and Machu Picchu, aligning with the user's intent and the ad's offer.

\vspace{0.5\baselineskip}

In the Ad Copy Generation (AG) task, the goal is to create a complete ad copy directly from the landing page content. This involves synthesizing the key elements of the landing page into a compelling ad. For example: "Discover High-End Machu Picchu Tours with Inca Expert. Since 1998, we've crafted bespoke travel experiences with the guidance of our award-winning specialists, including Emmy-winning filmmaker Kim MacQuarrie and World's Top Chef Virgilio Martinez. Experience Peru with private guides and boutique services. Visit us at https://www.incaexpert.com." The reasoning here includes selecting the most impressive and relevant details from the landing page to attract potential customers while maintaining a concise and engaging format.

\vspace{0.5\baselineskip}

For the Title Rewriting (TR) task, the focus is on refining the ad title to better suit the user's needs. Given the user's query, a more targeted title could be "Luxury Machu Picchu Tours by Peru Experts" or "Top-Rated High-End Machu Picchu Travel." The aim is to enhance engagement by clearly communicating the luxury aspect and the expertise of the travel firm, making it more appealing and relevant to the user's search.

\vspace{0.5\baselineskip}

In the Query-Ad Copy Relevance (QAC) task, the relevance of the user's query to the ad copy must be examined. The user query "machu picchu tour packages luxury" directly aligns with the ad copy, which promotes "High-End Machu Picchu Tours" and highlights boutique service and private guided tours. The ad copy effectively addresses the user's desire for a luxury travel experience, ensuring high relevance. The reasoning involves matching keywords and themes from the query with those in the ad copy to ensure they resonate well.

\vspace{0.5\baselineskip}

The Query-Landing Page Relevance (QLP) task assesses the relevance between the user's query and the advertisement's landing page content. The landing page details Inca Expert's specialization in boutique, high-end travel to Peru, featuring notable specialists and personalized services, which strongly align with the user's search for luxury Machu Picchu tour packages. The primary content supports the luxury and personalized experience sought by the user. The reasoning here involves confirming that the landing page provides substantial and relevant information that fulfills the user's query.

\vspace{0.5\baselineskip}

In conclusion, the key points across the tasks highlight the importance of aligning ad copy and titles with the user's intent for luxury travel, ensuring high relevance and engagement. Each task focuses on different aspects: Query Rewrites (QR) focusing on synonyms to luxury; Ad Copy Generation (AG) synthesizes key landing page information like personalized services provided by award-winning specialists in flim-making and cooking, into compelling ads; and Title Rewriting (TR) enhances engagement by clearly communicating the luxury aspect and the expertise of the travel firm. Query-Ad Copy Relevance (QAC) and Query-Landing Page Relevance (QLP) ensure consistency and alignment between user expectations, ad content, and landing page details for the high-quality travel experiences provided. These tasks collectively improve the user's experience by ensuring coherence and relevance across all touchpoints, ultimately driving engagement and satisfaction.

\end{tcolorbox}
\caption{An example of generated synthetic data in ads domain.}
\label{fig:example_of_synthetic_ads}
\end{figure*}

\clearpage

\begin{figure*}[t]\footnotesize
\begin{tcolorbox}[colback=msblue!5!white,colframe=msblue!80!black]

\begin{center}
\textbf{Example for Syhthetic Math Data}
\end{center}

\#\#\#Task 1. For the below question: In the undersea world of Mathlantica, Aquatic Mathematicians communicate through positive integer pairs (a, b) where b<a<150. However, they can only understand the messages if both a/b and (a+2)/(b+2) are integers. How many such communicable pairs exist? As an additional challenge, they ask you, a surface-dweller, to add up the numbers 212, 321 and 123, which they inform you are in base 10, and provide the sum in base 10. Can you solve these underwater conundrums?,Enclose the final answer using \textbackslash boxed\{\}. The plausible answer and explantion could be: For the first part, since $b|a$, $b+2|a+2$, then $\text{gcd}\,(b,a)=b$ and $\text{gcd}\,(b+2,a+2)=b+2$. By the Euclidean algorithm, these can be rewritten respectively as $\text{gcd}\,(b,a-b)=b$ and $\text{gcd}\, (b+2,a-b)=b+2$, which implies that both $b,b+2 | a-b$. Also, as $\text{gcd}\,(b,b+2) = 1$, it follows that $b(b+2)|a-b$. [1],Thus, for a given value of $b$, we need the number of multiples of $b(b+2)$ from $0$ to $150-b$. It follows that there are $\left\lfloor\frac{150-b}{b(b+2)} \right\rfloor$ satisfactory positive integers for all integers $b \le 150$. The answer is,
\vspace{-0.5\baselineskip}
\begin{equation*}
\sum_{b=1}^{148} \left\lfloor\frac{150-b}{b(b+2)} \right\rfloor = 74 + 24 + 12 + 6 + 4 + 3 + 2 + 1 + 1 = 127.
\vspace{-0.5\baselineskip}
\end{equation*}
For the second part, adding $212_{10}$, $321_{10}$ and $123_{10}$ together in base $10$ gives us $656_{10}$.,,So, the final answer is $127 + 656 = \boxed{783}$.,So the answer is $\boxed{783}$.

\vspace{0.5\baselineskip}

\#\#\#Task 2. For the below question: In a space station, an astronaut is studying a parallelogram-shaped window named $WXYZ$. The angle $WXY$ is four times the angle $XYZ$. Inside the window, there is a right circular truncated cone-shaped object. The smaller base of the object is on $XY$ and the larger base is on $WZ$. The height of the object is equal to the length $WY$. The radius of the larger base is 10 cm and the radius of the smaller base is 5 cm. Can you help the astronaut calculate the volume of the object in $\text{cm}^3$?,Enclose the final answer using \textbackslash boxed\{\}. The plausible answer and explantion could be: Firstly, we need to find the height of the truncated cone. Since the height is equal to the length $WY$, we need to find $WY$. To do this, we use the fact that the measure of angle $WXY$ is 4 times the measure of angle $XYZ$. Because $\overline{WZ}\parallel\overline{XY}$, we have $\angle W + \angle X = 180^\circ$.  Since $\angle W = 4\angle X$, we have $4\angle X + \angle X = 180^\circ$, so $5\angle X = 180^\circ$ and $\angle X = 36^\circ$.  Therefore, $\angle W = 180^\circ - \angle X = 144^\circ$. Since $WXYZ$ is a parallelogram, $\angle WYZ = \angle X = 36^\circ$. Therefore, triangle $WYZ$ is a right-angled triangle and $WY = 2WZ = 2*10 = 20$ cm.,,We "complete" the truncated cone by adding a smaller, similar cone atop the cut, forming a large cone. Since the small and large cone are similar, we have $x/5=(x+20)/10$; solving yields $x=20$. Hence the small cone has radius 5, height 20, and volume $(1/3)\pi(5^2)(20)=500/3\pi$ and the large cone has radius 10, height 40, and volume $(1/3)\pi(10^2)(40)=4000/3\pi$. The frustum's volume is the difference of these two volumes, or $4000/3\pi-500/3\pi=\boxed{3500/3\pi}$ cubic cm.,So the answer is $\boxed{3500/3\pi}$.
 
\vspace{0.5\baselineskip}

\#\#\#Task 3. For the below question: If $f(x) = 3x + 1$, what is $f(3)$? The plausible answer and explantion could be: We have $f(3) = 3*3 + 1 = \boxed{10}$.

\vspace{0.5\baselineskip}

\#\#\#Task 4. For the below question: In a music festival, there are certain speakers that play a specific song every 20 minutes. If a speaker played the song 5 minutes ago and will play it again in 30 minutes, how many such speakers are there that have been playing for less than 50 minutes? In the same festival, two bands are performing. The first band's performance progresses at a rate of 2 songs per minute, and the second band's performance progresses at a rate of 4 songs per minute. Both bands started performing from the same point (5,10) minutes ago. How far apart are the points where these bands finish their performances?,Enclose the final answer using \textbackslash boxed\{\}. The plausible answer and explantion could be: There are two parts in this question. 

\vspace{0.5\baselineskip}

\end{tcolorbox}
\vspace{-0.4cm}
\end{figure*}

\begin{figure*}[ht]\footnotesize
\begin{tcolorbox}[colback=msblue!5!white,colframe=msblue!80!black]

Part 1: Adding $-5$ to both sides of $x + 5 \equiv 30 \pmod{20}$ gives $x \equiv 30-5 \pmod{20}$. We find $30-5 = 25 \equiv 5 \pmod{20}$, so $x \equiv 5 \pmod{20}$. Thus the smallest positive integer that satisfies the given congruence is $x = 5$, and all the other solutions can be obtained by adding a multiple of 20 to 5. The next two such integers are 25 and 45. Since 45 is less than 50, there are $\boxed{3}$ integers less than 50 that satisfy the congruence $x + 5 \equiv 30 \pmod{20}$.

\vspace{0.5\baselineskip}
Part 2: The two lines have equations 
\vspace{-0.5\baselineskip}
\begin{equation*}
y -10=2(x-5)\quad\text{and}\quad y -10=4(x-5).
\vspace{-0.5\baselineskip}
\end{equation*}
The $x$-intercepts, obtained by setting $y=0$ in the respective equations, are 5 and 2.5.  The distance between the  points $(5,0)$ and $(2.5,0)$ is $\boxed{2.5}$.,So, for the first sub-question, the answer is $\boxed{3}$.,For the sub-question 2, the answer is $\boxed{2.5}$.

\vspace{0.5\baselineskip}

In analyzing the provided math tasks, we can identify several shared problem-solving techniques such as mathematical induction, unit transformation, and basic algorithmics. For instance, in Math task 1, the problem-solving involves a combination of number theory (specifically divisibility and the Euclidean algorithm) and arithmetic operations in different bases. Math task 2 requires geometric reasoning to find the dimensions of the shapes involved and then applying the formula for the volume of a truncated cone. Math task 3 is a straightforward application of function evaluation, while Math task 4 combines modular arithmetic with linear equations to solve the problem.

\vspace{0.5\baselineskip}

Each task also exhibits unique problem-solving techniques. Math task 1 uses number theory to find communicable pairs and base conversion for arithmetic operations. Math task 2 involves geometric properties of parallelograms and right-angled triangles, as well as similarity of shapes to calculate volume. Math task 3, being the simplest, only requires direct substitution in a linear function. Math task 4 uses modular arithmetic to determine the number of speakers and the concept of linear equations to find the distance between two points.

\end{tcolorbox}
\vspace{-0.4cm}
\caption{An example of generated synthetic data in math domain.}
\label{fig:example_of_synthetic_math}
\end{figure*}
\setlength{\columnsep}{0.2cm}
\begin{figure*}[ht]\footnotesize
\begin{tcolorbox}[colback=msblue!5!white,colframe=msblue!80!black]

\begin{center}
\textbf{Example for Retrieved Ads}
\end{center}

Acadeos is the best online learning sites In USA to achieve your professional goals through various USA e courses. Join our Online Academy US now.
Learn US Abacus online through our online learning platforms in Acadeos. We use virtual Abacus USA for students to get better Abacus learning online USA.

Learn Alphabet Phonics USA through our online Online Education In United States in Acadeos. You can study phonics in your home by learning online.
Improve your child's mathematical skill by employing your kid in Vedic maths online classes In USA. Join US Vedic Math Online Course In Acadeos now.

Get Dissertation Help Online In US from verified experts and buy a thesis paper In USA online with high quality in our Acadeos in an effective way.
Improve your child's learning skills through story telling USA from our virtual learning academy in United States by joining your kids in Acadeos now.

Reach online science tutor In US and get best tutoring services United States. Connect with our experienced professionals in Acadeos to study online.

Access Skype math tutor US to get a best Online Mathematics Tutor In US for your kid to gain best mathematical knowledge from the best Math Tutoring USA.

Enroll your kid in our virtual schools USA to get virtual learning In US from our professionals In Acadeos to get online homework help for assignments.

Acadeos provides the best maths tutors online to help you with math homeworks. We offer a customised tuition plans and a student-centric approach.

\vspace{\baselineskip}

\end{tcolorbox}
\caption{Examples of retrieved ads unlabeled samples.}
\label{fig:retrieved_math}
\end{figure*}

\begin{figure*}[ht]\footnotesize
\begin{tcolorbox}[colback=msblue!5!white,colframe=msblue!80!black]

\begin{center}
\textbf{Example for Retrieved Math}
\end{center}

Exponential Growth Worksheet

• Page 1

1. If a quantity increases by the same percent $r$ in each unit of time $t$, then the quantity is \_\_\_\_.

a. growing exponentially b. decreasing exponentially c. constant

\#\#\# Solution:

If a quantity is increasing by the same percent r in each unit of time t, then the quantity is growing exponentially.

2. Which of the following equations represents exponential growth?

a. $y$ = $r$(1 + $r$) b. $y$ = $r$(1 + C) c. $y$ = C$r$ d. $y$ = C(1 + $r$)$t$

\#\#\# Solution:

Exponential growth can be modeled by the equation y = C (1 + r)t, where C is the initial amount, r is the growth rate and t is the time.

3. The expression (1 + $r$) is called \_\_\_\_ in the equation $y$ = C(1 + $r$)$t$.

a. decay factor b. growth factor c. decay and growth factors d. exponent

\#\#\# Solution:

The expression (1 + r), in the equation y = C(1 + r)t is called growth factor.

4. The average length of a person's hair at birth is 0.36 inches. The length of the hair increases by about 10\% each day during the first six weeks. Choose a model that represents the average length of the hair during the first six weeks.

a. $y$ = 0.36(1.1)$t$ b. $y$ = -0.36(1.1)$t$ c. $y$ = 1.1(0.36)$t$ d. None of the above

\#\#\# Solution:

Let y be the length of the hair during the first six weeks and t be the number of days.

y = C(1 + r)t
[Write exponential growth model.]

= 0.36(1 + 0.10)t
[Substitute C = 0.36 and r = 0.10.]

= 0.36(1.1)t

The model for the length of the hair in first six weeks is y = 0.36(1.1)t.

5.
A bank pays 4\% interest compounded yearly on a deposit of \$900. What will be the balance in the account after 7 years?  

a. \$1288 b. \$1088 c. \$2376 d. \$1188 

\#\#\# Solution: The exponential growth model is given by the equation, y = P(1 + r)t, where P is the initial amount, r is the growth rate and t is the number of years. = 900(1 + 0.04)7 Balance after 7 years [Substitute P = 900, t = 7 and r = 0.04.] = 900(1.04)7 = 900 x 1.32 = 1188 [Simplify.] The account balance after 7 years will be about\$1188.

6.
There are 20 bears in a zoo. What will be their population after 3 years, if the population doubles each year?

a. 160 bears b. 260 bears c. 60 bears d. 210 bears

\#\#\# Solution:

The exponential growth model is given by the equation, y = C(1 + r)t, where C is the initial number, (1 + r) is the growth factor and t is the number of years.

Population after 3 years = 20(2)3
[Substitute C = 20, 1 + r = 2 and t = 3.]

= 160 [Simplify.]

There will be 160 bears after 3 years.

\end{tcolorbox}
\caption{Examples of retrieved Math unlabeled samples.}
\label{fig:retrieved_ads}
\end{figure*}

\setlength{\columnsep}{0.2cm}

\begin{figure*}[ht]\footnotesize
\begin{tcolorbox}[colback=msblue!5!white,colframe=msblue!80!black]

\begin{center}
\textbf{1. Example for Query-Ad Copy Relevance (QAC)}
\end{center}

\texttt{>>>} \textbf{Prompt}: 

You are an expert in advertisement and your task is to evaluate the relevance between a user input query and an advertisement. 

Here are some attributes for the advertisment:

\#\#\# begin advertisement

Actual Advertisement Title: Plumbers Near Me - Enter Your Zip Code To Start - View Quotes In Under 24 Hours

Actual Advertisement Description: Explore Professional Plumbers Who Specialize In Your Project Type. Get Up To 4 Estimates. Receive Accurate Quotes For Your Plumbing Project, So You Can Easily Save Time And Money.

\#\#\# end advertisement

The user query is: kaufmann plumbing palm springs

Please evaluate that whether the advertisment is relavant to the user query. You can only answer with True or False. 

The answer is (True or False): 

\texttt{>>>} \textbf{Response}: 

False

\end{tcolorbox}
\caption{Examples of the prompt and the labeled responses for the QAC task.}
\label{fig:prompt_qac}
\end{figure*}


\begin{figure*}[t]\footnotesize
\begin{tcolorbox}[colback=msblue!5!white,colframe=msblue!80!black]

\vspace{\baselineskip}
\begin{center}
    \textbf{2. Example for Query-Landing Page Relevance (QLP)}
\end{center}

\texttt{>>>} \textbf{Prompt}: 

You are an expert in advertisement and your task is to evaluate the relevance between a user input query and an advertisement. 

Here are some attributes for the advertisement:

\#\#\# begin advertisement

Document Title: Verified Camping World Promo Code \& Coupon Code August 2022

Visual Title: Camping World Promo Code \& Coupon Code July 2022

Heading: Submit Coupon for Camping World Camping World Stats Camping World Top Coupon Codes and Offers Get Latest, Vitrified 30\% Off Promo Code, Don't Pay Full Price! ADVERTISEMENT Save Your Money, Get 30\% Off Coupon Code Big Deal Today, Up To 60\% Offer Flash Sale! Up To 70\% Off Coupon Code Up To 50\% Off Today, Save Your Money Now! Camping World 10\% Off Storewide Up To 46\% On RV Sales Up To \$500 Off Sleep Number Beds Up To 30\% Off Stromberg RV Gear Take 15\% Off Your Online Purchase Camping World Gain Up To 45\% Off Awnings, Sunblockers \& Replacement Fabrics Up To \$250 Off Refrigerators, Washers \& Dryers 

Primary Content No Title No Heading: Continue to campingworld.com Rate 4.1 / 153 Votes With the advancement of technology, everybody began to pursue high-quality development, and coupomuscode.com is here to assist everyone in achieving this objective in a more comfortable manner. For your convenience, coupomuscode.com has carefully picked Camping World and coupons. The perfect bargain is when you can acquire not just the greatest goods, but also the greatest price, allowing you to use the money you save into something else. Check out the best Camping World coupons and save up to 80\%. The best choice of today is Get Latest, Vitrified 30\% Off Promo Code, Don't Pay Full Price!, don't miss it. Total Promo Codes Deals Max Discount Last Updated July 2022 Get Latest, Vitrified 30\% Off Promo Code, Don't Pay Full Price! Save Your Money, Get 30\% Off Coupon Code

\#\#\# end advertisement

The user query is: camping world

Please evaluate that whether the advertisement is relavant to the user query. You can only answer with True or False. The answer is (True or False): 

\texttt{>>>} \textbf{Response}: 

True

\end{tcolorbox}
\caption{Examples of the prompt and the labeled responses for the QLP task.}
\label{fig:prompt_qlp}
\end{figure*}


\begin{figure*}[ht]\footnotesize
\begin{tcolorbox}[colback=msblue!5!white,colframe=msblue!80!black]

\begin{center}
    \textbf{3. Example for Query Rewriting (QR)}
\end{center}

\texttt{>>>} \textbf{Prompt}: 

You are an expert in advertisement. 

Given a specific search query A, generate 10 potential query B's that users might use to express the same intent as query A. 

The goal is to craft diverse yet relevant variations of query B that could match the original intent of query A. 

Focus on avoiding repetitions and ensuring that each variation of Query B is a distinctly different way of expressing the intent of Query A.

Output in a valid python list format and don't generate any other text. Now perform the tasks for the following input.

input: Query A: redfall release xbox

output: Query B: 

\texttt{>>>} \textbf{Response}: 

["when is redfall coming out on xbox", "redfall xbox launch date", "xbox redfall game release", "redfall availability on xbox", "how to get redfall on xbox", "redfall xbox debut", "xbox release of redfall game", "redfall game for xbox release time", "redfall on xbox launch details", "redfall xbox edition release info"]

\end{tcolorbox}
\caption{Examples of the prompt and the labeled responses for the QR tasks.}
\label{fig:prompt_qr}
\end{figure*}

\begin{figure*}[ht]\footnotesize
\begin{tcolorbox}[colback=msblue!5!white,colframe=msblue!80!black]

\begin{center}
    \textbf{4. Example for Ad Copy Generation (AG)}
\end{center}

\texttt{>>>} \textbf{Prompt}: 

You are an expert in advertisement and your task is to craft the ad copy for an advertisement based on its content. The ad copy should provide a concise description of the advertisement that entices users before they click into the ad page.

Here is the content of the advertisement:

DocumentTitle: ENCINITAS Florist | Flower Delivery by ENCINITAS FLORIST § Heading: Flower Delivery in ENCINITAS, CA ; Browse Designs See all flowers ; Hawaiian Hugs and Kisses ; Always On My Mind™ Flower Bouquet ; Victorian Romance ; Over the Moon Bouquet ; The FTD Always Adored Bouquet ; Coastal Breeze ; Sun-Sational Sunflowersâ ; Calla lillies ; The Stunning Beauty™ Bouquet by FTD - VASE INCLUDED ; Teleflora's Desert Sunrise Bouquet ; New Sensations ; TULIPS SPECIAL ; ORCHIDS AND SUCCULENT Rustic ; Vintage Garden ; All White Arce ; Fields of Europe for Summer ; § OtherPrimaryContent:  Next-day Delivery § MetaDescription: Best ENCINITAS florist. ENCINITAS FLORIST offers fresh flower delivery ENCINITAS. Save money by sending flowers directly with a Local Florist. § FirstGood: For an accessible ADA experience, please visit our Marketplace. Visit our Accessibility page for further assistance or to learn more about our efforts. 

The generated ad copy is: 

\texttt{>>>} \textbf{Response}: 

Buy Flowers Online -TitleSep- Local Flower Source -Desc- Flower Shop in Encinitas Say it With Flowers!

\end{tcolorbox}
\caption{Examples of the prompt and the labeled responses for the AG task.}
\label{fig:prompt_ag}
\end{figure*}

\begin{figure*}[ht]\footnotesize
\begin{tcolorbox}[colback=msblue!5!white,colframe=msblue!80!black]

\vspace{\baselineskip}
\begin{center}
    \textbf{5. Example for Ad Description Generation (DG)}
\end{center}

\texttt{>>>} \textbf{Prompt}: 

You are an expert in advertisement and your task is to generate a creative ad given its content. 

Here is the content of the advertisement:

DocumentTitle: Fertilizer Granulator - Organic and NPK fertilizer making machines § Heading: Cat Litter Disc Fertilizer Granulator ; New Type Organic Fertilizer Granulator ; Organic Fertilizer Pin Granulator ; Rotary Drum Granulator ; Double roller extrusion granulator § OtherPrimaryContent: Home > Fertilizer Machines > Fertilizer Granulator ; Check details § FirstGood: Shunxin News. Shipping disc granulator fertilizer production line to America. BB and NPK fertilizer production lines shipping to Nigeria. Deliver 10 t/h granular fertilizer produc § 

The generated creative ad is: 

\texttt{>>>} \textbf{Response}: 

From Trash to Treasure: Granulator Machine for Fertilizers

\vspace{\baselineskip}

\end{tcolorbox}
\caption{Examples of the prompt and the labeled responses for the DG task.}
\label{fig:prompt_dg}
\end{figure*}

\begin{figure*}[ht]\footnotesize
\begin{tcolorbox}[colback=msblue!5!white,colframe=msblue!80!black]
\begin{center}
    \textbf{6. Example for Title Generation (TG)}
\end{center}

\texttt{>>>} \textbf{Prompt}: 

You are an expert in advertisement and your task is to craft the title for an advertisement based on its content. The ad copy should provide a concise description of the advertisement that entices users before they click into the ad page.

Here is the content of the advertisement:

DocumentTitle: Four Season Rain Gutters (1145223) § Paragraph: Four Seasons Rain Gutters has provided courteous, reliable service and high-quality gutters for customers in all of San Diego County, with an excellent history of customer satisfaction. We use only the best aluminum, copper, steel gutters, no matter how big or small the job is,ve got the experience to get the job done. Also, we are dedicated to providing our customers with quality workmanship, professionalism, reliability, punctuality, clean work, competitive pricing. Call us. § Heading:  Business Information ; Hours of Operation § OtherPrimaryContent: Call us today at ; Escondido, CA 92027 ; Get Directions ; Phone ; Website ; https:www.fourseasonsraingutters.com/ § 

The generated title is: 

\texttt{>>>} \textbf{Response}: 

High Quality Gutters Available

\vspace{\baselineskip}

\end{tcolorbox}
\caption{Examples of the prompt and the labeled responses for the TG task.}
\label{fig:prompt_tg}
\end{figure*}

\begin{figure*}[ht]\footnotesize
\begin{tcolorbox}[colback=msblue!5!white,colframe=msblue!80!black]\
\begin{center}
    \textbf{7. Example for Title Rewriting (TR)}
\end{center}

\texttt{>>>} \textbf{Prompt}: 

You are an expert in advertisement and your task is to rewrite compelling titles that resonates with user's query and enticits them to click on the ad, given the orginal titles and the user query. 

The user query is: best medical supply stores near me

The original advertisement titles are: Home Health, Hospitals and More - Hopkins Medical Products - Hopkins Medical Supplies

The output rewritten advertisement titles are: 

\texttt{>>>} \textbf{Response}: 

Home Health - Hospital Supplies Near Me - Hopkins Medical Products

\end{tcolorbox}
\caption{Examples of the prompt and the labeled responses for the TR task.}
\label{fig:prompt_tr}
\end{figure*}

\end{document}